%% file: ms.tex
\documentclass{article}

\input{math_commands.tex}

\usepackage[utf8]{inputenc}
\pdfoutput=1
\usepackage[sc,osf]{mathpazo}
\usepackage[margin=1.1in,letterpaper]{geometry}
\usepackage[sort&compress,numbers]{natbib}
\usepackage[colorlinks=true,citecolor=blue,breaklinks]{hyperref}
\usepackage[hyphenbreaks]{breakurl} 
\usepackage{caption}
\usepackage{subcaption}

\makeatletter
\newlength\aftertitskip     \newlength\beforetitskip
\newlength\interauthorskip  \newlength\aftermaketitskip

\def\maketitle{\par
 \begingroup
   \def\thefootnote{\fnsymbol{footnote}}
   \def\@makefnmark{\hbox to 4pt{$^{\@thefnmark}$\hss}}
   \@maketitle \@thanks
 \endgroup
\setcounter{footnote}{0}
 \let\maketitle\relax \let\@maketitle\relax
 \gdef\@thanks{}\gdef\@author{}\gdef\@title{}\let\thanks\relax}

\def\@startauthor{\noindent \normalsize\bf}
\def\@endauthor{}
\def\@starteditor{\noindent \small {\bf Editor:~}}
\def\@endeditor{\normalsize}
\def\@maketitle{\vbox{\hsize\textwidth
 \linewidth\hsize \vskip \beforetitskip
 {\begin{center} \LARGE\@title \par \end{center}} \vskip \aftertitskip
 {\def\and{\unskip\enspace{\rm and}\enspace}%
  \def\addr{\small\it}%
  \def\email{\hfill\small\tt}%
  \def\name{\normalsize\bf}%
  \def\AND{\@endauthor\rm\hss \vskip \interauthorskip \@startauthor}
  \@startauthor \@author \@endauthor}
}}

\makeatother

\pdfoutput=1                    % force pdflatex to run

\usepackage[utf8]{inputenc} % allow utf-8 input
\usepackage[T1]{fontenc}    % use 8-bit T1 fonts
\usepackage{hyperref}       % hyperlinks

\usepackage{url}            % simple URL typesetting
\usepackage{booktabs}       % professional-quality tables
\usepackage{amsfonts}       % blackboard math symbols
\usepackage{nicefrac}       % compact symbols for 1/2, etc.
\usepackage{microtype}      % microtypography
\usepackage[utf8]{inputenc}

\usepackage{graphicx}
\usepackage{framed}
\usepackage{amssymb}
\usepackage{amsfonts}
\usepackage{mathrsfs}
\usepackage{mathtools}
\usepackage{array}
\usepackage{amsthm}
\usepackage{verbatim} 
\usepackage{enumerate}
\usepackage{bbm}
\usepackage{mathtools}
\usepackage{commath}
\usepackage{wrapfig}
\usepackage{amsbsy}
\usepackage{subcaption}
\usepackage{float}

\usepackage{url}
\usepackage{amsmath}
\usepackage{amsthm}
\usepackage{algorithm}
\usepackage{mathtools}
\usepackage{amssymb}
\usepackage{tikz}
\usepackage[noend]{algpseudocode}
\usepackage{wrapfig}
\usepackage{hyperref}

\usepackage{algorithm}
\usepackage[noend]{algpseudocode}

\title{The Role of Embedding Complexity in \\ Domain-invariant Representations}

\author{\name Ching-Yao Chuang \email{cychuang@mit.edu}\\
  \name Antonio Torralba \email{torralba@mit.edu}\\
  \name Stefanie Jegelka \email{stefje@mit.edu}\\
  \addr{CSAIL, Massachusetts Institute of Technology, Cambridge, MA 02139} 
}

\begin{document}

\maketitle

\begin{abstract}
  Unsupervised domain adaptation aims to generalize the hypothesis trained in a source domain to an unlabeled target domain. One popular approach to this problem is to learn domain-invariant embeddings for both domains. In this work, we study, theoretically and empirically, the effect of the embedding complexity on generalization to the target domain. In particular, this complexity affects an upper bound on the target risk; this is reflected in experiments, too. Next, we specify our theoretical framework to multilayer neural networks. As a result, we develop a strategy that mitigates sensitivity to the embedding complexity, and empirically achieves performance on par with or better than the best layer-dependent complexity tradeoff.

\end{abstract}

\section{Introduction}
\label{intro}

%[Intro to Unsupervised Domain Adaptation]
Domain adaptation is critical in many applications where collecting large-scale supervised data is prohibitively expensive or intractable, or where conditions at prediction time can change. For instance, self-driving cars must be robust to different weather, change of landscape and traffic. In such cases, the model learned from limited source data should ideally generalize to different target domains. Specifically, unsupervised domain adaptation aims to transfer knowledge learned from a labeled source domain to similar but completely unlabeled target domains. 

%[Intro to Domain Invariant Representation]
One popular approach to unsupervised domain adaptation is to learn domain-invariant representations \citep{ben2007analysis,long2015learning,ganin2016domain}, by minimizing a 
divergence between the representations of source and target domains. The prediction function is learned on these ``aligned'' representations with the aim of making it domain-independent. A series of theoretical works justifies this idea \citep{ben2007analysis,mansour2009domain,ben2010theory,cortes2011domain}.

%[Problems recent discovered of domain invariant representation]
Despite the empirical success of domain-invariant representations, exactly matching the representations of source and target distribution can sometimes fail to achieve domain adaptation. For example, \citet{wu2019domain} show that exact matching may increase target error if label distributions are different between source and target domain, and propose a new divergence metric to overcome this limitation.
\citet{zhao2019learning} establish lower and upper bounds on the risk when label distributions between source and target domains differ.
\citet{johansson2019support} point out the information lost in non-invertible embeddings, and propose different generalization bounds based on the overlap of the supports of source and target distribution.

In contrast to previous analyses that focus on changes in the label distributions or joint support, we study the effect of embedding complexity. In particular, we show a general bound on the target risk that reflects
a tradeoff between embedding complexity and the divergence of source and target domains. A too powerful class of embeddings can result in overfitting the source data and the matching of source and target distributions, resulting in arbitrarily high target risk. Hence, a restriction %(taking into account assumptions about correspondences and invariances)
is needed. We observe that indeed, without appropriately constraining the embedding complexity, the performance of state-of-the-art methods such as domain-adversarial neural networks \citep{ganin2016domain} can deteriorate significantly. 

Next, we tailor the bound to multilayer neural networks. In a realistic scenario, one may have a total depth budget and divide the network into an encoder (embedding) and predictor by aligning the representations of source and target in a chosen layer, which defines the division. In this case, a more complex encoder necessarily implies a weaker predictor, and vice versa. This tradeoff is reflected in the bound and, we see that, in practice, there is an ``optimal" division.

To better optimize the tradeoff between encoder and predictor without having to tune the division, we propose to
optimize the tradeoffs in all layers jointly via a simple yet effective objective that can easily be combined with most current approaches for learning domain-invariant representations. Implicitly, this objective restricts the more powerful deeper encoders by encouraging a simultaneous alignment across layers. In practice, the resulting algorithm achieves performance on par with or better than standard domain-invariant representations, without tuning of the division.

Empirically, we examine our theory and learning algorithms on sentiment analysis (Amazon review dataset), digit classification (MNIST, MNIST-M, SVHN) and general object classification (Office-31). In short, this work makes the following contributions:
\begin{itemize}
\item General upper bounds on target error that capture the effect of embedding complexity when learning domain-invariant representations;
\item Fine-grained analysis for multilayer neural networks, and a new objective with implicit regularization that stabilizes and improves performance;
\item Empirical validation of the analyzed tradeoffs and proposed algorithm on several datasets.
\end{itemize}

\section{Unsupervised Domain Adaptation}
\label{sec_uda}

For simplicity of exposition, we consider binary classification with input space $\mathcal X \subseteq \mathbb{R}^{n}$ and output space $\mathcal Y = \{0, 1\}$. Define $\mathcal H$ to be the hypothesis class from $\mathcal X$ to $\mathcal Y$. The learning algorithm obtains two datasets: labeled source data $\mathcal X_{S}$ from distribution $p_{S}$, and unlabeled target data $\mathcal X_{T}$ from distribution $p_{T}$. We will use $p_S$ and $p_T$ to denote the joint distribution on data and labels $X,Y$ and the marginals, i.e., $p_S(X)$ and $p_S(Y)$. Unsupervised domain adaptation seeks a hypothesis $h \in \mathcal H$ that minimizes the risk in the target domain measured by a loss function $\ell$ (here, zero-one loss):
\begin{align}
R_{T}(h) = \mathbb{E}_{x,y \sim p_{T}}[\ell(h(x), y)].
\label{eq:rt}
\end{align}
%here we consider standard L2 loss for discrete labels in this work. Analogously to (\ref{eq:rt}), we define the source risk as $R_{S}(h) = \mathbb{E}_{x,y \sim p_{S}}[\ell(h(x), y)]$.  Note that
We will not assume common support in source and target domain, in line with standard benchmarks for domain adaptation such as adapting from MNIST to MNIST-M.

\subsection{Domain-invariant Representations}
A common approach to domain adaptation is to learn a joint embedding of source and target data \citep{ganin2016domain,tzeng2017adversarial}. The idea is that aligning source and target distributions in a latent space $\mathcal Z$ results in a domain-invariant representations, and hence a subsequent classifier $f$ from the embedding to $\mathcal Y$ will generalize from source to target. Formally, this results in the following objective function on the hypothesis $h = fg := f \circ g$, where $\mathcal{G}$ is the class of embedding functions from $\mathcal{X}$ to $\mathcal{Z}$, and we minimize a divergence $d$ between the distributions $p_{S}^{g}(Z) = p_{S}(g(X)), p_T^{g}(Z)= p_{T}(g(X))$ of source and target after mapping to $\mathcal{Z}$:

\begin{align}
\min_{f \in \mathcal{F}, g \in \mathcal{G}} R_{S}(fg) + \alpha d(p_{S}^{g}(Z), p_{T}^{g}(Z)).
\label{di_obj}
\end{align}
The divergence $d$ could be, e.g., the Jensen-Shannon \citep{ganin2016domain} or Wasserstein distance \citep{shen2017wasserstein}.

\subsection{Upper bounds on the target risk}
\citet{ben2007analysis} introduced the $\mathcal{H} \Delta \mathcal{H}$-divergence to bound the worst-case loss from extrapolating between domains. Let $R_{D}(h, h^{\prime}) = \mathbb{E}_{x \sim D}[\ell(h(x), h^{\prime}(x))]$ be the
expected disagreement between two hypotheses. The $\mathcal{H} \Delta \mathcal{H}$-divergence measures whether there is any pair of hypotheses whose disagreement (risk) differs a lot between source and target distribution.

\newtheorem{thm}{Theorem}

\newtheorem{definition}[thm]{Definition}
\begin{definition}
\textnormal{($\mathcal{H}\Delta \mathcal{H}$-divergence)} Given two domain distributions $p_{S}$ and $p_{T}$ over $\mathcal{X}$, and a hypothesis class $\mathcal{H}$, the $\mathcal{H}\Delta \mathcal{H}$-divergence between $p_{S}$ and $p_{T}$ is
\begin{align*}
d_{\mathcal{H}\Delta \mathcal{H}}(p_{S}, p_{T}) = \sup_{h, h^{\prime} \in \mathcal{H}} |R_{S}(h, h^{\prime}) - R_{T}(h, h^{\prime})|.
\end{align*}
\end{definition}
The $\mathcal{H}\Delta \mathcal{H}$-divergence is determined by the discrepancy between source and target distribution and the complexity ofthe hypothesis class $\mathcal{H}$.
For a hypothesis class $\mathcal{H}: \mathcal{X} \rightarrow \{0,1\}$, the disagreement between two hypotheses is equivalent to the exclusive or function. Hence, one can interpret the $\mathcal{H}\Delta \mathcal{H}$-divergence as finding a classifier in function space $\mathcal{H}\Delta \mathcal{H} = \mathcal{H}\oplus\mathcal{H}$ which attempts to maximally separate one domain from the other \citep{ben2010theory}. A restrictive hypothesis space may result in small $\mathcal{H}\Delta \mathcal{H}$-divergence even if the source and target domain do not share common support. This divergence allows us to bound the risk on the target domain:

\newtheorem{theorem}[thm]{Theorem}
%\numberwithin{theorem}{section}
\begin{theorem}\label{thm:bendavid}
  \textnormal{\citep{ben2010theory}} %Assume the setting in section \ref{sec_uda},
  For all hypotheses $h \in \mathcal H$, the target risk is bounded as
\begin{align*}
R_{T}(h) \leq R_{S}(h) + d_{\mathcal{H}\Delta \mathcal{H}}(p_{S}, p_{T}) + \lambda_{\mathcal H},
\end{align*}
where $\lambda_{\mathcal H}$ is the best joint risk 
\begin{align*}
\lambda_{\mathcal H} \coloneqq \inf_{h' \in \mathcal{H}}[R_S(h') + R_T(h')]
\end{align*}
\end{theorem} 

Similar results exist for continuous labels \citep{cortes2011domain,mansour2009domain}. 

Theorem \ref{thm:bendavid} is an influential theoretical result in unsupervised domain adaptation, and motivated work on domain invariant representations. For example, recent work (\citet{ganin2016domain,johansson2019support}) applied Theorem \ref{thm:bendavid} to the hypothesis space $\mathcal F$ that maps the representation space $\mathcal{Z}$ induced by an encoder $g$ to the output space:
\begin{align}
R_{T}(fg) \leq R_{S}(fg) + d_{\mathcal F \Delta \mathcal F}(p_{S}^{g}(Z), p_{T}^{g}(Z)) + \lambda_{\mathcal F}(g)
\label{bound1}
\end{align}
where $\lambda_{\mathcal F}(g)$ is the best hypothesis risk with fixed $g$, i.e., $\lambda_{\mathcal F}(g) \coloneqq \inf_{f' \in \mathcal{F}}[R_S(f'g) + R_T(f'g)]$. The $\mathcal F \Delta \mathcal F$ divergence implicitly depends on the fixed $g$ and can be small if $g$ provides a suitable representation. However, if $g$ induces a wrong alignment, then the best hypothesis risk $\lambda_{\mathcal F}(g)$ is large with any function class $\mathcal{F}$. The following example will illustrate such a situation, motivating to explicitly take the class of embeddings into account when bounding the target risk.

%%%%%%%%%%%%%%%%%%%
%   SECTION 3
%%%%%%%%%%%%%%%%%%%
\section{Influence of the embedding complexity}

\begin{figure}
  \begin{minipage}[c]{0.6\textwidth}
    \includegraphics[width=\linewidth]{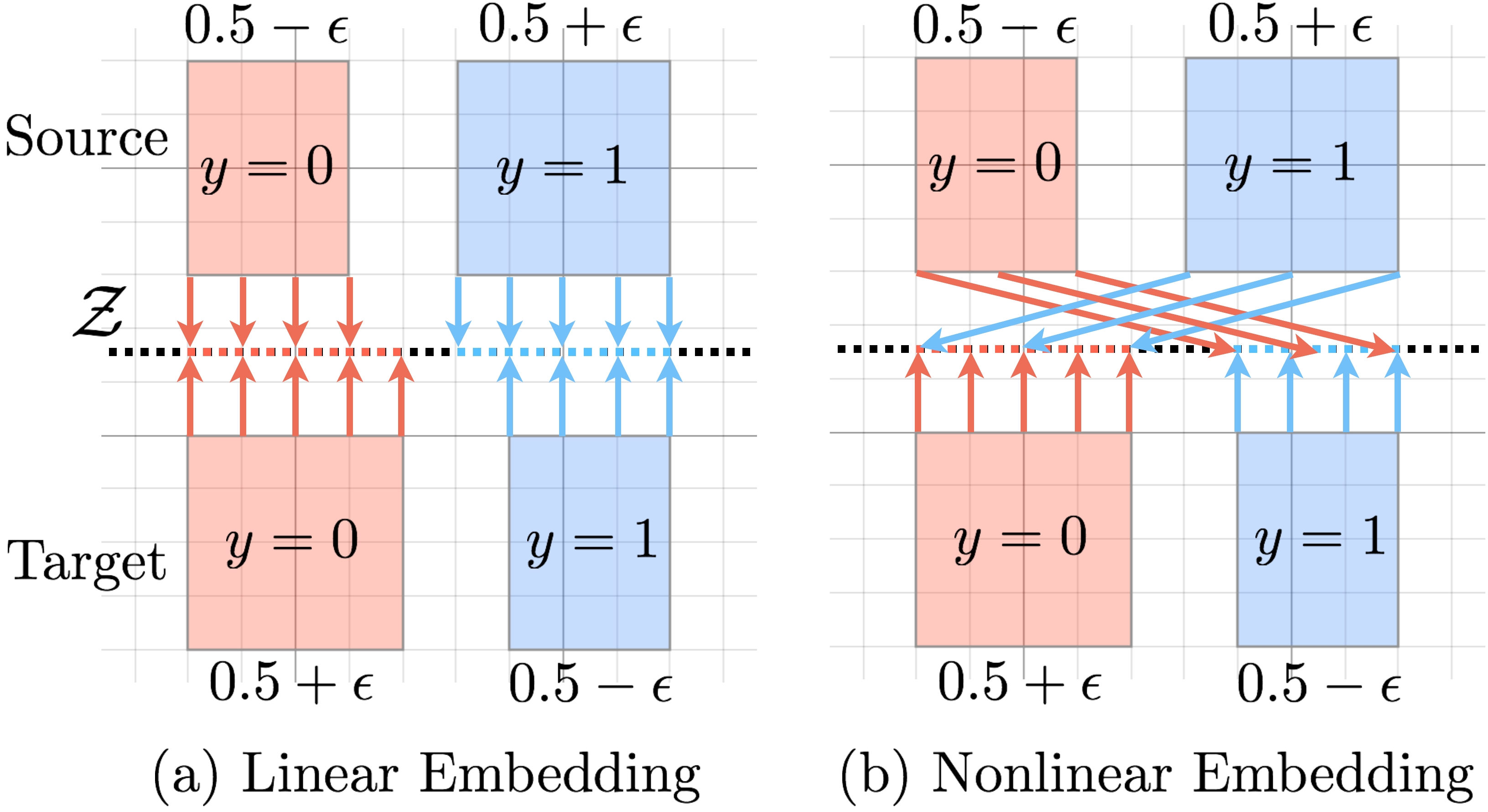}
  \end{minipage}\hfill
  \begin{minipage}[c]{0.35\textwidth}
    \vspace{-0mm}
    \caption{
       Illustrative example in 2D. The 1D representation space is illustrated as a dotted line, and arrows indicate the embedding from 2D to 1D. (a) Optimal embedding when $\mathcal{G}$ is the class of linear functions. % from 2D to 1D.
    (b) Optimal embedding with a complex nonlinear function class: zero source error and divergence loss, but the embedding destroys label consistency and leads to maximal target risk.
    } \label{fig_2d}
  \end{minipage}
  \vspace{-1mm}
\end{figure}

%\subsection{A Motivating example}
We begin with an illustrative toy example. Figure~\ref{fig_2d} shows a binary classification problem in 2D with disjoint support and a slight shift in the label distributions from source to target: $p_{S}(y=1) = p_{T}(y=1) + 2\epsilon$. Assume the representation space $\mathcal{Z}$ is one dimensional, so the embedding $g$ is a function from 2D to 1D. If we allow arbitrary, nonlinear embeddings, then, for instance, the embedding in Figure~\ref{fig_2d}(b), together with an optimal predictor, achieves zero source loss and a zero divergence which is optimal according to the objective in \eqref{di_obj}. But the target risk of this combination of embedding and predictor is maximal: $R_T(fg) = 1$.

If we restrict the class $\mathcal{G}$ of embeddings to linear maps $g(x) = \mathbf{W}x$ where $\mathbf{W} \in {\rm I\!R}^{1 \times 2}$, then the embeddings that are optimal with respect to the objective \eqreff{di_obj} are of the form $\mathbf{W} = \begin{bmatrix}
    a ,  0 
\end{bmatrix}$.
Together with an optimal source classifier $f$, they achieve a non-zero value of $2\epsilon$ for objective \eqreff{di_obj} due to the shift in class distributions. However, these embeddings retain label correspondences and can thus minimize target risk.

This example illustrates that a too rich class of embeddings can ``overfit'' the alignment, and hence lead to arbitrarily bad solutions. Hence, the complexity of the encoder class plays an important role in learning domain invariant representations.

\subsection{Bounds for Domain-invariant Representations}
Motivated by the above example, we next expose how the bound on the target risk depends on the complexity of the embedding class. To do so, we apply Theorem \ref{thm:bendavid} to the hypothesis $h = fg$:

\begin{align}
%\vspace{-3mm}
R_{T}(fg) \leq R_{S}(fg) + d_{\mathcal F \mathcal{G} \Delta \mathcal F \mathcal{G}}(p_{S}, p_{T}) + \lambda_{\mathcal F \mathcal{G}}.
\label{bound2}
%\vspace{-3mm}
\end{align}
This bound differs in two ways from the previous bound (\eqref{bound1}), which was based only on $\mathcal{F}$: the best in-class joint risk now minimizes over both $\mathcal{F}$ and $\mathcal{G}$, i.e.,
\begin{align}
  \label{eq:lambdafg}
  \lambda_{\mathcal F \mathcal G} \coloneqq \inf_{f \in \mathcal{F}, g \in \mathcal{G}}[R_S(fg) + R_T(fg)],
\end{align}
  which is smaller than $\lambda_{\mathcal F}(g)$ and reflects the fact that we are learning both $f$ and $g$. In return, the divergence term $d_{\mathcal F \mathcal{G} \Delta \mathcal F \mathcal{G}}(p_{S}, p_{T})$ becomes larger than the one in \eqref{bound1}. To better understand these tradeoffs, we will reformulate bound \eqreff{bound2} to be more interpretable. To this end, we define a version of the $\mathcal{H}\Delta\mathcal{H}$-divergence that explicitly measures variation of the embeddings in $\mathcal{G}$:
  
\begin{definition}
\textnormal{($\mathcal{F}_{\mathcal{G}\Delta \mathcal{G}}$-divergence)} For two domain distributions $p_{S}$ and $p_{T}$ over $\mathcal{X}$, an encoder class $\mathcal{G}$, and predictor class $\mathcal{F}$, the $\mathcal{F}_{\mathcal{G}\Delta \mathcal{G}}$-divergence between $p_{S}$ and $p_{T}$ is
\begin{align*}
      d_{\mathcal{F}_{\mathcal{G}\Delta\mathcal{G}}}(p_{S}, p_{T}) = \sup_{f \in \mathcal{F}; \; g,g^{\prime} \in \mathcal{G}} |R_{S}(fg, fg^{\prime}) - R_{T}(fg, fg^{\prime})|.
\end{align*}
\label{ordef}
\vspace{-3mm}
\end{definition}  

Importantly, the $\mathcal{F}_{\mathcal{G}\Delta \mathcal{G}}$-divergence is smaller than the $\mathcal{FG}\Delta \mathcal{FG}$-divergence, since the two hypotheses in the supremum, $fg$ and $fg^{\prime}$, share the same predictor $f$. 
\begin{theorem}\label{thm:main}
  %Given the setting in section \ref{sec_uda},
  For all $f \in \mathcal F$ and $g \in \mathcal G$,
\begin{align}
R_{T}(fg) \leq R_{S}(fg) + \underbrace{d_{\mathcal{F}\Delta\mathcal{F}}(p_{S}^{g}(Z), p_{T}^{g}(Z))}_{\text{\textnormal{Latent Divergence}}} + \underbrace{d_{\mathcal{F}_{\mathcal{G}\Delta\mathcal{G}}}(p_{S}, p_{T})}_{\textnormal{Embedding Complexity}}+\lambda_{\mathcal F\mathcal{G}}(g).
\end{align}
where $\lambda_{\mathcal F\mathcal{G}}(g)$ is the best in-class joint risk defined as
\begin{align}
    \lambda_{\mathcal F\mathcal{G}}(g) =  \inf_{f^{\prime} \in \mathcal{F}, g^{\prime} \in \mathcal{G}} 2R_{S}(f^{\prime}g) + R_{S}(f^{\prime}g^{\prime}) + R_{T}(f^{\prime}g^{\prime}). \nonumber
\end{align}
\end{theorem}

We prove all theoretical results in the Appendix. %Proofs of all Propositions and Theorems
%can be found in the Appendix.
This target generalization bound is small if (C1) the source risk is small, (C2) the latent divergence is small (because the domains are well-aligned and/or $\mathcal{F}$ is restricted), (C3) the complexity of $\mathcal{G}$ is restricted to avoid overfitting of alignments, and (C4) good source and target risk is in general achievable with $\mathcal{F}$ and $\mathcal{G}$.

\paragraph{Comparison to Previous Bounds.}
The last two terms in Theorem \ref{thm:bendavid} express a similar complexity tradeoff, but with respect to the overall hypothesis class $\mathcal{H}$, which here combines encoder and predictor. Directly applying Theorem~\ref{thm:bendavid} to the composition $\mathcal{H} = \mathcal{F}\mathcal{G}$ (\eqref{bound2}) treats both jointly and does not make the role of the embedding as explicit as Theorem~\ref{thm:main}.

The recent bound~\eqreff{bound1} assumes a fixed embedding $g$ and focuses on the predictor class $\mathcal{F}$. As a result, it captures embedding complexity even less explicitly: the first two terms in bound \eqreff{bound1} and Theorem~\ref{thm:main} are the same. The last term in \eqreff{bound1}, $\lambda_{\mathcal{F}}(g)$, contains the target risk with the given $g$. Hence, bound \eqreff{bound1} replaces (C3) and (C4) above by saying $\mathcal{F}$ and the specific $g$ (which is much harder to control since in practice it is also optimized) can achieve good source and target risk. In contrast, Theorem~\ref{thm:main} states an explicit complexity penalty on the variability of the embeddings, and uses the fixed $g$ only in the source risk, which can be better estimated empirically.

If $\mathcal{F}$ is not too rich, the latent divergence can be empirically minimized by finding a well-aligned embedding. Hence, we can minimize the upper bound in Theorem~\ref{thm:main} by minimizing the usual source loss and domain-invariant loss~\eqreff{di_obj} and by choosing $\mathcal{F}$ and $\mathcal{G}$ appropriately to tradeoff the complexity penalty $d_{\mathcal{F}_{\mathcal{G}\Delta \mathcal{G}}}$, the latent divergence (which increases with complexity of $\mathcal{F}$ and decreases with complexity of $\mathcal{G}$), and the best in-class joint risk (which decreases with complexity of $\mathcal{F}$ and $\mathcal{G}$).

\subsection{Embedding Complexity Tradeoffs Empirically}
\label{sec_33}

\iffalse
In this section, we unveil the role of embedding complexity in learning domain-invariant representations by investigating the encoder class $\mathcal{G}$ in the generalization bound. The source error $R_{S}(fg)$ and the latent divergence $d_{\mathcal{F}\Delta\mathcal{F}}(p_{S}(Z_g), p_{T}(Z_g))$ are the terms we can empirically minimized in the right hand side of the bound. The $d_{\mathcal{F}_{\mathcal{G}\Delta\mathcal{G}}}(p_{S}, p_{T})$ is determined by the function class $\mathcal{F}$ and $\mathcal{G}$ and can not be empirically minimized. It becomes larger as the embedding complexity increases. Therefore, it can be seen as the \textit{complexity penalty} of minimizing the domain-invariant loss. Without restricting the complexity of the encoder class $\mathcal{G}$, the complexity penalty can be large, indicating that the target risk may be large too. However, a less complex encoder class can increase the latent divergence since the encoder might not have sufficient representation power to minimize the domain divergence. Therefore, latent divergence and input divergence complexity penalty makes a trade-off explicit between the divergence and the encoder complexity. 
\fi

\begin{figure}
  \begin{minipage}[c]{0.68\textwidth}
    \includegraphics[width=\linewidth]{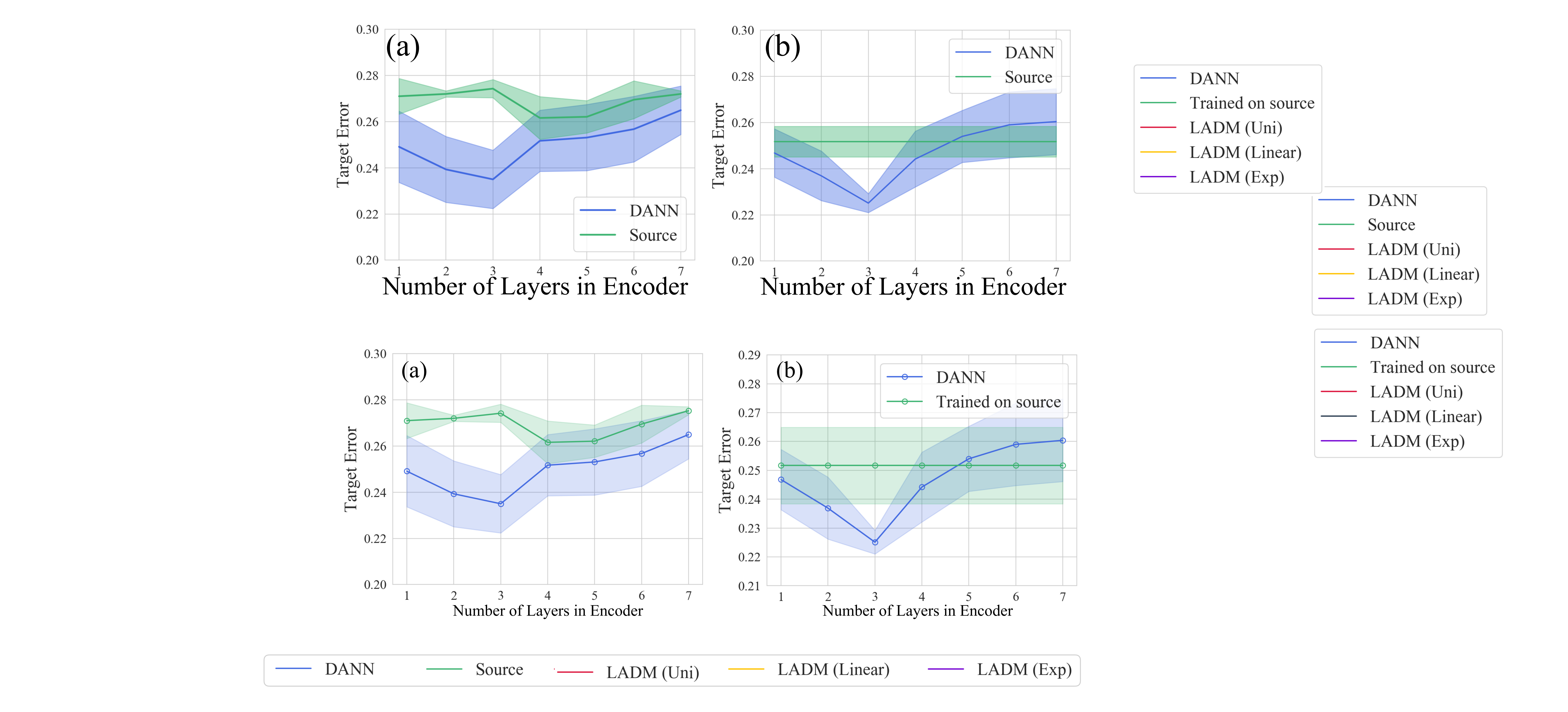}
  \end{minipage}\hfill
  \begin{minipage}[c]{0.29\textwidth}
    \caption{
       Empirical verification on Amazon reviews dataset. (a) Vary the number of layers in the encoder while fixing the predictor. (b) Fix the total number of layers and optimize the domain-invariant loss in different layers.
    } \label{fig_ev}
  \end{minipage}
\end{figure}

To empirically verify the embedding complexity tradeoff, we keep the predictor class $\mathcal{F}$ fixed, vary the embedding class $\mathcal{G}$, and minimize the source loss and alignment objective~\eqreff{di_obj}. Concretely, we 
train domain adversarial neural networks (DANNs) \citep{ganin2016domain} on the Amazon reviews dataset (Book $\rightarrow$ Kitchen). Our hypothesis class is a multi-layer ReLU network, and the divergence is minimized against a discriminator. For more experimental details and results, please refer to section \ref{exp}. We train different models by varying the number of layers in the encoder while fixing the predictor to $4$ layers. Figure~\ref{fig_ev}(a) shows that, when increasing the number of layers in the encoder, the target error decreases initially and then increases as more layers are added. This supports our theory: the smaller encoders are not rich enough to allow for good alignments and $\lambda_{\mathcal{FG}}(g)$, but overly expressive encoders may overfit.

\textbf{Predictor Complexity.} Theoretically, the complexity of the predictor class $\mathcal{F}$ also affects the generalization bound in Theorem~\ref{thm:main}. Empirically, we found that the predictor complexity has much weaker influence on the target risk (see experiments in Appendix~\ref{a_pred_trad}). Indeed, theoretically, while the complexity of $\mathcal{F}$ affects the latent divergence, if the alignment via $g$ is very good, this divergence can still be small. In addition, the $\mathcal{F}_{\mathcal{G}\Delta \mathcal{G}}$-divergence is more sensitive to the embedding complexity than the predictor complexity. This offers a possible explanation for our observations. In the remainder of this paper, we focus on the role of the embedding.

\iffalse
\subsubsection{Remark on Predictor Complexity}
The complexity of the predictor will also affect the bound. When we increase the complexity of the predictor, both the latent space and input space divergences will increase. If the best in-class risk $\lambda_{\mathcal{FG}(g)}$ is already small, increasing the predictor complexity will only lead to worse performance. Essentially, there is no trade-off between the divergences and the predictor complexity. Besides, the $\mathcal{F}_{\mathcal{G}\Delta \mathcal{G}}$-divergence is less sensitive to the predictor class comparing to the encoder class by definition: $fg$ and $fg^{\prime}$ in the supremum of Definition \ref{ordef} share the same predictor but different encoders. Complex predictor does not necessarily lead to a large $\mathcal{F}_{\mathcal{G}\Delta \mathcal{G}}$-divergence if the encoder class is not powerful enough. Therefore, we will focus on embedding complexity trade-off in the main paper. More discussion and experiments about predictor complexity can be found in the appendix \ref{a_pred_trad}.
\fi

\textbf{Discussion.} The results in this section indicate that, 
without constraining the embedding complexity, we may overfit the distribution alignment and thereby destroy label consistency as in Figure~\ref{fig_2d}. The bound suggests to choose the minimal complexity encoder class $\mathcal{G}$ that is is still expressive enough to minimize the latent space divergence. Practically, this can be done by regularizing the encoder, e.g., restricting Lipschitz constants or norms of weight matrices. More explicitly, one may limit the number of layers of a neural network, or
apply inductive biases via network architectures. For instance, compared to fully connected networks, convolutional neural networks (CNNs) restrict the output representations to be spatially consistent with respect to the input.

\iffalse
% COMMENT COMMENT COMMENT COMMENT
To reduce the worst case divergence \circled{i}, we need to restrict the encoder family to those that can approximately minimize \circled{i}, in coordination with the predictor class $\mathcal{F}$. Practically, we can optimize the original objective of domain invariant representations in Equation~\ref{di_obj} to align the latent distributions.

Term \circled{ii} implies that we should choose the minimal complexity encoder class $\mathcal G$ that is is still expressive enough to encode the data from both domains. Practically, this can be done by regularizing the encoder, e.g., restricting Lipschitz constants or norms of weight matrices. More explicitly, one may limit the number of layers of a neural network, or apply inductive biases via selecting network architectures. For instance, comparing to fully connected networks (FCs), convolutional neural networks (CNNs) restrict the output to be spatially consistent with respect to the input.
% COMMENT COMMENT COMMENT COMMENT
\fi 
\vspace{-1mm}
\section{Bounds for Multilayer Neural Networks}
\vspace{-1mm}
\label{sec_bo}
Due to their wide empirical success, multilayer neural networks 
%Multilayer neural networks have achieved significant success across various tasks and
have been adopted for learning domain-invariant representations. Next, we adapt the bound in Theorem \ref{thm:main} to multilayer networks. Specifically, we consider the number of layers as an explicit measurement of complexity.
%and discuss the corresponding trade-offs. Through the lens of our theoretical finding, we propose
This will lead to a simple yet effective algorithm to mitigate the negative effect of very rich encoders.

\subsection{Effect of Layer Divisions}
\vspace{-1.5mm}
Assume we have an \textit{$N$-layer feedforward neural network} $h \in \mathcal{H}$. The model 
%
%Due to the decomposability of multilayer neural networks,
$h$ can be decomposed as $h = f_{i}g_{i} \in \mathcal{F}_{i}\mathcal{G}_{i} = \mathcal{H}$ for $i \in \{1, 2, \dots, N-1 \}$ where the embedding $g_{i}$ is formed by the first layer to the $i$-th layer and the predictor $f_{i}$ is formed by the $i+1$-th layer to the last layer. We can then rewrite the bound in Theorem \ref{thm:main} in layer-specific form:
\begin{align}
\label{eq_lsb}
R_{T}(h) \leq R_{S}(h) + \underbrace{d_{\mathcal{F}_{i}\Delta\mathcal{F}_{i}}(p_{S}^{g_{i}}(Z), p_{T}^{g_{i}}(Z))}_{\substack{\text{\textnormal{Latent Divergence}}  \text{\textnormal{ in $i$-th layer}} }} + \underbrace{d_{{\mathcal{F}_{i}}_{\mathcal{G}_{i}\Delta\mathcal{G}_{i}}}(p_{S}, p_{T})}_{\substack{\text{\textnormal{Embedding Complexity}}  \text{\textnormal{ w.r.t $\mathcal{G}_i$}} }}+\lambda_{\mathcal F_{i}\mathcal{G}_{i}}(g_{i}).
\end{align}
This yields $N-1$ layer-specific upper bounds.

Importantly, minimizing the domain-invariant loss in different layers leads to different tradeoffs between fit and complexity penalties. This is reflected by the following inequalities that relate different layer divisions.

\newtheorem{proposition}[thm]{Proposition}
%\numberwithin{corollary}{section}
\begin{proposition}\label{prop}
\textnormal{\textbf{(Monotonicity)}}
In an $N$-layer feedforward neural network $h = f_{i}g_{i} \in \mathcal{F}_{i}\mathcal{G}_{i} = \mathcal{H}$ for $i \in \{1, 2, \dots, N-1 \}$, the following inequalities hold for all $i \leq j$:
\begin{align}
    d_{{\mathcal{F}_{i}}_{\mathcal{G}_{i}\Delta\mathcal{G}_{i}}}(p_{S}, p_{T}) &\leq
    d_{{\mathcal{F}_{j}}_{\mathcal{G}_{j}\Delta\mathcal{G}_{j}}}(p_{S}, p_{T})  &\text{\text(embedding complexity)} \label{ineq1}\\
         d_{{\mathcal{F}_{i}}\Delta\mathcal{F}_{i}}(p_{S}^{g_{i}}(Z), p_{T}^{g_{i}}(Z)) &\geq
    d_{{\mathcal{F}_{j}}\Delta\mathcal{F}_{j}}(p_{S}^{g_{j}}(Z), p_{T}^{g_{j}}(Z)) &\text{(latent divergence)} \label{ineq2}
\end{align}
\end{proposition} 
Proposition \ref{prop} states that the latent divergence is monotonically decreasing and the complexity penalty is monotonically increasing with respect to the embedding's depth. This is a tradeoff within the fixed combined hypothesis class $\mathcal{H}$.
%The above inequalities form a trade-off between the complexity of encoder and predictor even the complexity of the hypothesis is fixed.
A deeper embedding allows for better alignments and simultaneously reduces the depth (power) of $\mathcal{F}$; both reduce the latent divergence. At the same time, it incurs a larger $\mathcal{F}_{\mathcal{G}\Delta\mathcal{G}}$-divergence.

This suggests that there might be an optimal division that minimizes the bound on the target risk. In practice, this translates into the question: \textit{in which intermediate layer should we optimize the domain-invariant loss?} Figure~\ref{fig_ev}(b) shows how the target error changes as a function of the layer division, with a total of $n=8$ layers. Indeed, empirically there is an optimal division with minimum target error, suggesting that for a fixed $\mathcal{H}$, i.e., total network depth, not all divisions are equal.

If the exact layer-specific bounds could be computed, one could simply select the layer division with the lowest bound. But, this is in general computationally nontrivial. Instead, we take a different perspective. In fact, the layer-specific bounds \eqreff{eq_lsb} all hold simultaneously, \emph{independent of the layer we selected for distribution alignment}.
\newtheorem{corollary}[thm]{Corollary}
\begin{corollary}\label{cor}
Let $h$ be an $N$-layer feedforward neural network $h = f_{i}g_{i} \in \mathcal{F}_{i}\mathcal{G}_{i} = \mathcal{H}$ for $i \in \{1, 2, \dots, N-1 \}$, we have the layer-agnostic bound
\begin{align*}
R_{T}(h) \leq R_{S}(h) + \min_{\{1 \leq i < N\}} \Big\{ d_{\mathcal{F}_{i}\Delta\mathcal{F}_{i}}(p_{S}^{g_{i}}(Z), p_{T}^{g_{i}}(Z)) + d_{{\mathcal{F}_{i}}_{\mathcal{G}_{i}\Delta\mathcal{G}_{i}}}(p_{S}, p_{T})+\lambda_{\mathcal F_{i}\mathcal{G}_{i}}(g_{i}) \Big\}.
\end{align*} %\label{eq_cor}
where $\lambda_{\mathcal F\mathcal{G}}(g)$ is the best in-class joint risk defined in Theorem \ref{thm:main}.
\end{corollary} 
The corollary implies that at least one of these bounds should be small. Recall that the bounds depend on how well we can minimize the source risk and align the distributions via a sufficiently powerful embedding, while, at the same time, limiting the complexity of $\mathcal{F}$ and $\mathcal{G}$.

\subsection{Multilayer Divergence Minimization and Regularization}
Corollary~\ref{cor} points to various algorithmic ideas: (1) \emph{Simultaneously} optimizing several bounds may result in approximately minimizing at least one of them, without having to select an optimal one. (2) We may attain small latent divergence with a deeper encoder, if we achieve to restrict the complexity of $\mathcal{G}$ appropriately. It turns out that these two ideas are related.

Optimizing the domain-invariant loss with alignment in a specific layer may result in large bounds for the other layers, due to the monotonicity of the two divergences (Proposition~\ref{prop}) and potentially non-aligned embeddings in lower layers. Hence, we propose to instead solve a multi-objective optimization problem where we \emph{jointly} align source and target distributions in multiple layers. Let $\mathcal{L} \subseteq \{1,2, \dots, N-1\}$ be a subset of layers. We minimize the weighted sum of divergences, and refer to this objective as  \emph{Multilayer Divergence Minimization (MDM)}:
\begin{align}
\min\nolimits_{h \in \mathcal{H}}\; R_{S}(h) + \sum\nolimits_{i \in \mathcal{L}} \alpha_{i} d(p_{S}^{g_{i}}(Z), p_{T}^{g_{i}}(Z)).
\label{la_obj}
\end{align}
This objective encourages alignment throughout the layer-wise embeddings in the network.
First, a good alignment minimizes the latent divergence, if $\mathcal{F}$ is not too rich. For the lower layers (shallow embeddings), this comes together with a very restricted class of embeddings, and hence limits both latent divergence and complexity penalty. Without the optimization across layers, the embeddings in lower layers are not driven towards alignment.

Second, enforcing alignment in lower layers implicitly restricts the deeper embeddings in higher layers, since the embeddings are such that alignment happens early on. This effect may be viewed as an implicit regularization. By this perspective, the bounds for higher layers profit from low latent divergences (deeper embeddings and shallow predictors) and restricted complexity of $\mathcal{G}$. 

In general, one can simply set $\mathcal{L} = \{1,2, \dots, N-1\}$. To improve computational efficiency, we can sub-sample layers or exclude the first and the last few layers. MDM is simple and general, and can be combined with most algorithms for learning domain-invariant representations. For DANN, for instance, we minimize the divergence in multiple layers by adding discriminators.

\vspace{-1.5mm}
\section{Other Related Works}
\vspace{-1.5mm}
Existing approaches for learning domain-invariant representations may be distinguised, e.g., by which divergence they measure between source and target domain. Examples include domain adversarial learning approaches \citep{ganin2014unsupervised, tzeng2015simultaneous, ganin2016domain}, maximum mean discrepancy (MMD) \citep{long2014transfer, long2015learning, long2016unsupervised} and Wasserstein distance \citep{courty2016optimal, courty2017joint, shen2017wasserstein, lee2018minimax}.

Other works improve performance by combining the domain-invariant loss with other objectives.
%Domain-invariant loss has also been combined with the other objectives to improve the performance.
\citet{shu2018dirt} penalize the violation of the cluster assumption. In addition to the shared feature encoder between source and target domain,
\citet{bousmalis2016domain} include private encoders for each domain to capture domain-specific information. \citet{long2018conditional} propose a domain discriminator that is conditioned on the cross-covariance of domain-specific embeddings and classifier predictions to leverage discriminative information.
Besides the usual distribution alignment,
\citet{hoffman2017cycada} further align the input space with a generative model that maps the target input distribution to the source distribution. These previous works can be interpreted as adding additional regularization via auxiliary objectives, and thereby potentially reducing the complexity penalty.

Some previous works also optimize the domain-invariant loss in multiple layers. \citet{long2016unsupervised} fuse the representations from a bottleneck layer and a classifier layer by a tensor product and minimize the domain divergence based on the aggregated representations. 
Joint adaptation networks (JADs) \citep{long2017deep} minimize the MMD in the last few layers %(e.g., last three layers of AlexNet \citep{krizhevsky2012imagenet} or last two layers of ResNet \citep{he2016deep})
to make the embeddings more transferable. MDM can be seen as a generalization of JADs that minimizes domain divergence in nearly every layer, % or the subset of it
driven by a strong theoretical motivation. Importantly, minimizing the divergence only in the last few layers could still be suboptimal, since the embeddings may not be sufficiently regularized. % according to our analysis.

\vspace{-3mm}
\section{Experiments}
\vspace{-3mm}
\label{exp}

\begin{figure}
    \includegraphics[width=\linewidth]{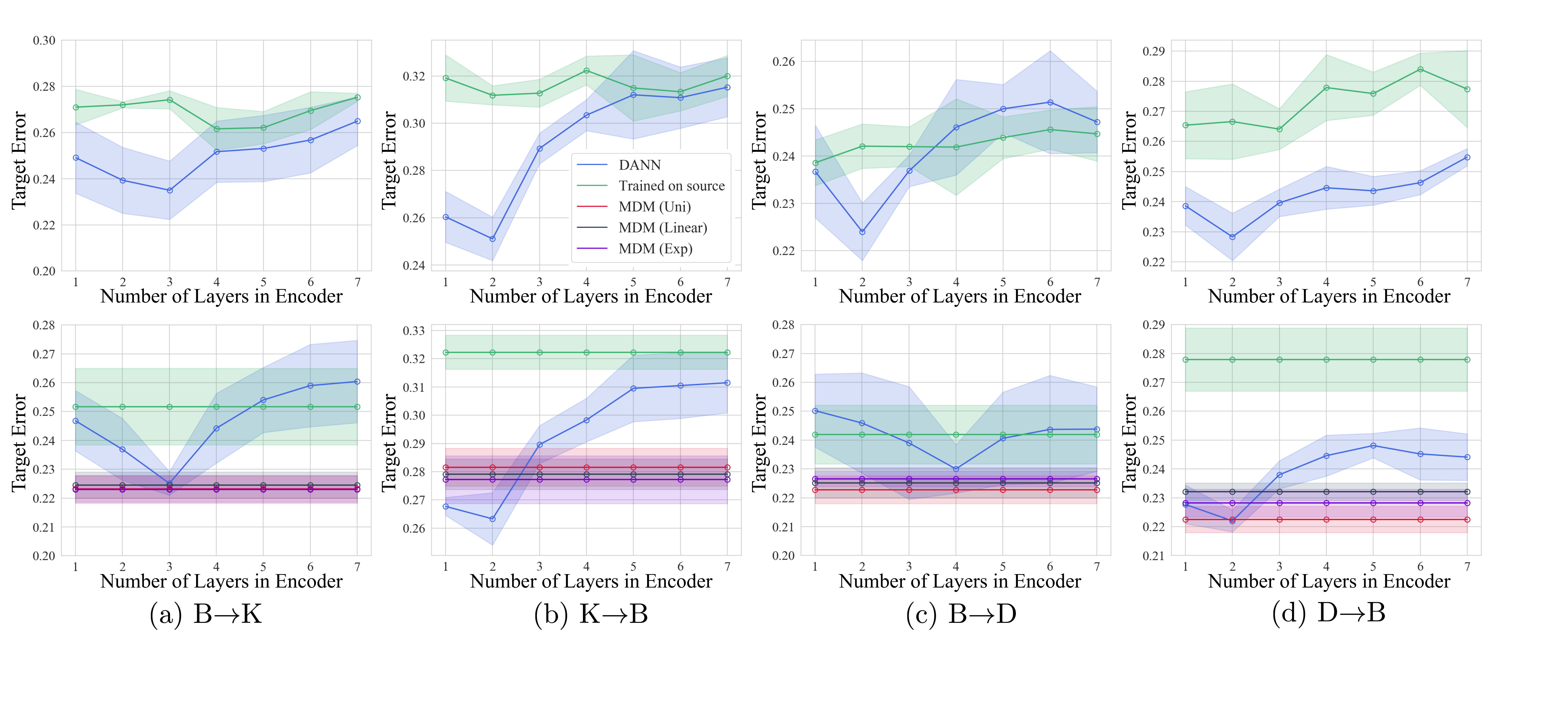}
    \vspace{-5mm}
    \caption{
       Amazon reviews dataset. First row: Fixed predictor class, varying number of layers in the encoder. Second row: Fixed total number of layers and optimizing domain-invariant loss in a single intermediate layer or MDM.
    } \label{fig_amazon}
  \vspace{-2mm}
\end{figure}

We test our theory and algorithm on several standard benchmarks: sentiment analysis (Amazon reviews dataset), digit classification (MNIST, MNIST-M, SVHN) and general object classification (Office-31). In all experiments, we train DANN \citep{ganin2016domain}, which measures the latent divergence via a domain discriminator (Jensen Shannon divergence).  A validation set from the source domain is used as an early stopping criterion during learning. In all experiments, we use the Adam optimizer \citep{kingma2014adam} and a progressive training strategy for the discriminator \citep{ganin2016domain}. We primarily consider three types of complexity: number of layers, number of hidden neurons, and inductive bias (CNNs). In all experiments, we retrain each model for 5 times and plot the mean and standard deviation of the target error.

For evaluating MDM\footnote{The code is available at \url{https://github.com/chingyaoc/mdm}}, we consider three weighting schemes: uniform weights ($\alpha_{i} = \alpha_{0}$), linearly decreasing ($\alpha_{i} = \alpha_{0} - c \times i$), and exponentially decreasing ($\alpha_{i} = \alpha_{0}\exp(-c \times i)$) where $c \geq 0$. The decreasing weights encourage the network to minimize the latent divergence in the first few layers, where the embedding complexity is low. This may also further restrict the deeper embeddings.
%
%since the corresponding complexity would be smaller (Proposition \ref{prop}) and together result in a smaller bound.
More experimental details can be found in Appendix \ref{a_arch}.

\paragraph{Sentiment Classification.}
We first examine complexity tradeoffs on the Amazon reviews data, which has four domains (books (B), DVD disks (D), electronics (E), and kitchen appliances (K)) with binary labels (positive / negative review). Reviews are encoded into 5000 dimensional feature vectors of unigrams and bigrams. The hypothesis class are multi-layer ReLU networks. We show the results on B$\rightarrow$K, K$\rightarrow$B, B$\rightarrow$D, and D$\rightarrow$B in Figure~\ref{fig_amazon}.
To probe the effect of embedding complexity by itself, we fix the predictor class to $4$ layers and vary the number of layers of the embedding.
%We plot the results in the upper row of Figure \ref{fig_amazon}.
In agreement with the results in Section~\ref{sec_33}, the target error decreases initially, and then increases as more layers are added to the encoder. 

Next, we probe the tradeoff when the total number of layers is fixed to $8$.
The bottom row of Figure~\ref{fig_amazon} shows that there exists an optimal setting for all tasks.
For MDM, we optimize alignment in all intermediate layers. The results suggest that MDM's performance is comparable to the hypothesis with the optimal division, without tuning the division. The three weighting schemes perform similarly, suggesting that MDM is robust to weight selection.

\paragraph{Digit Classification.}

\begin{figure}
    \includegraphics[width=\linewidth]{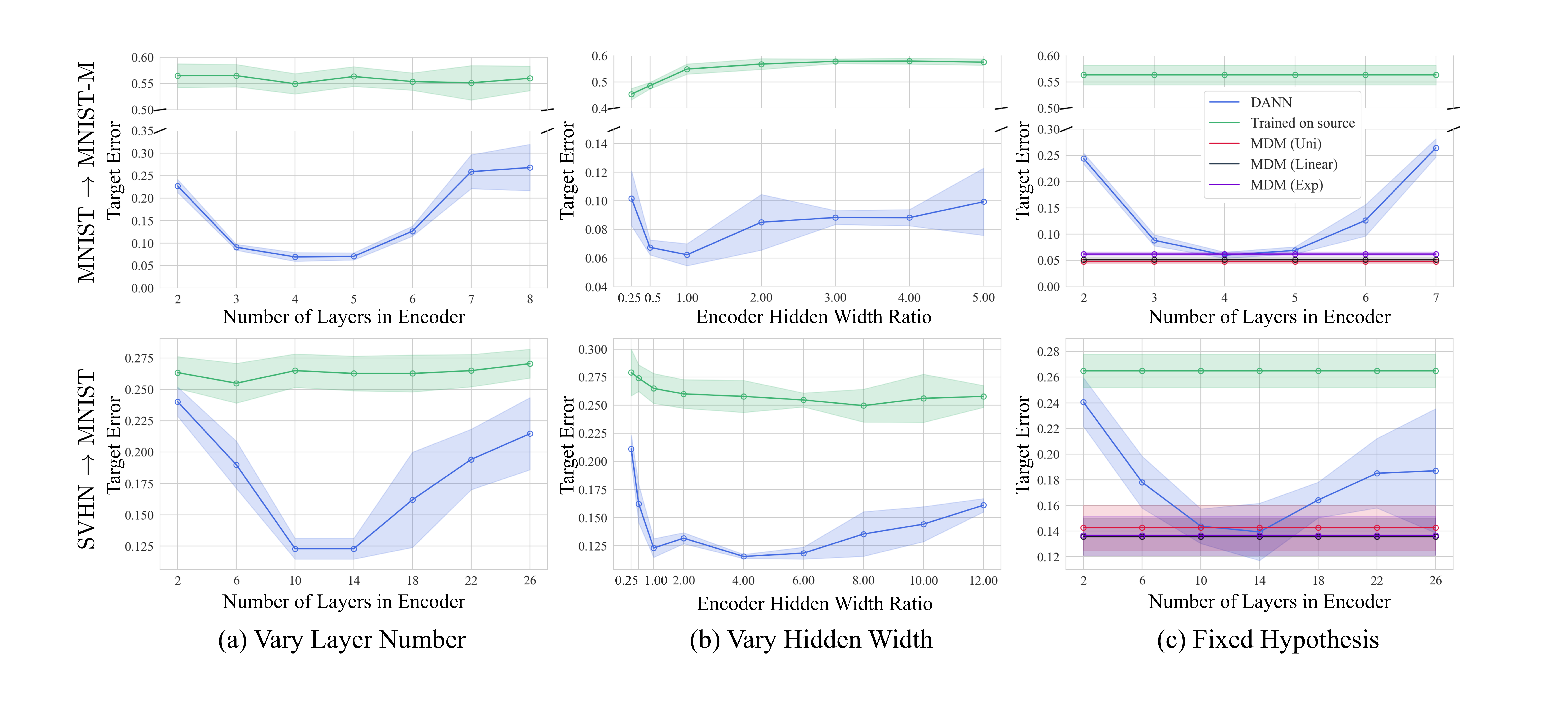}
    \vspace{-7mm}
    \caption{
       Digit classification. (a) Fixed predictor class, varying number of layers in the encoder. (b) Fixed predictor class, varying the hidden width of the encoder. (c) Fixed total number of layers and optimizing domain-invariant loss in a single intermediate layer or MDM.
    } \label{fig_mnist}
  \vspace{-0mm}
\end{figure}

We next verify our findings on standard  domain adaptation benchmarks: MNIST$\rightarrow$ MNIST-M (M$\rightarrow$M-M) and SVHN$\rightarrow$MNIST (S$\rightarrow$M). We use standard CNNs as the hypothesis class; architecture details are in Appendix \ref{a_arch}. 

\begin{wrapfigure}{r}{0.35\textwidth}
  \begin{center}
  \vspace{-9mm}
    \includegraphics[width=0.35\textwidth]{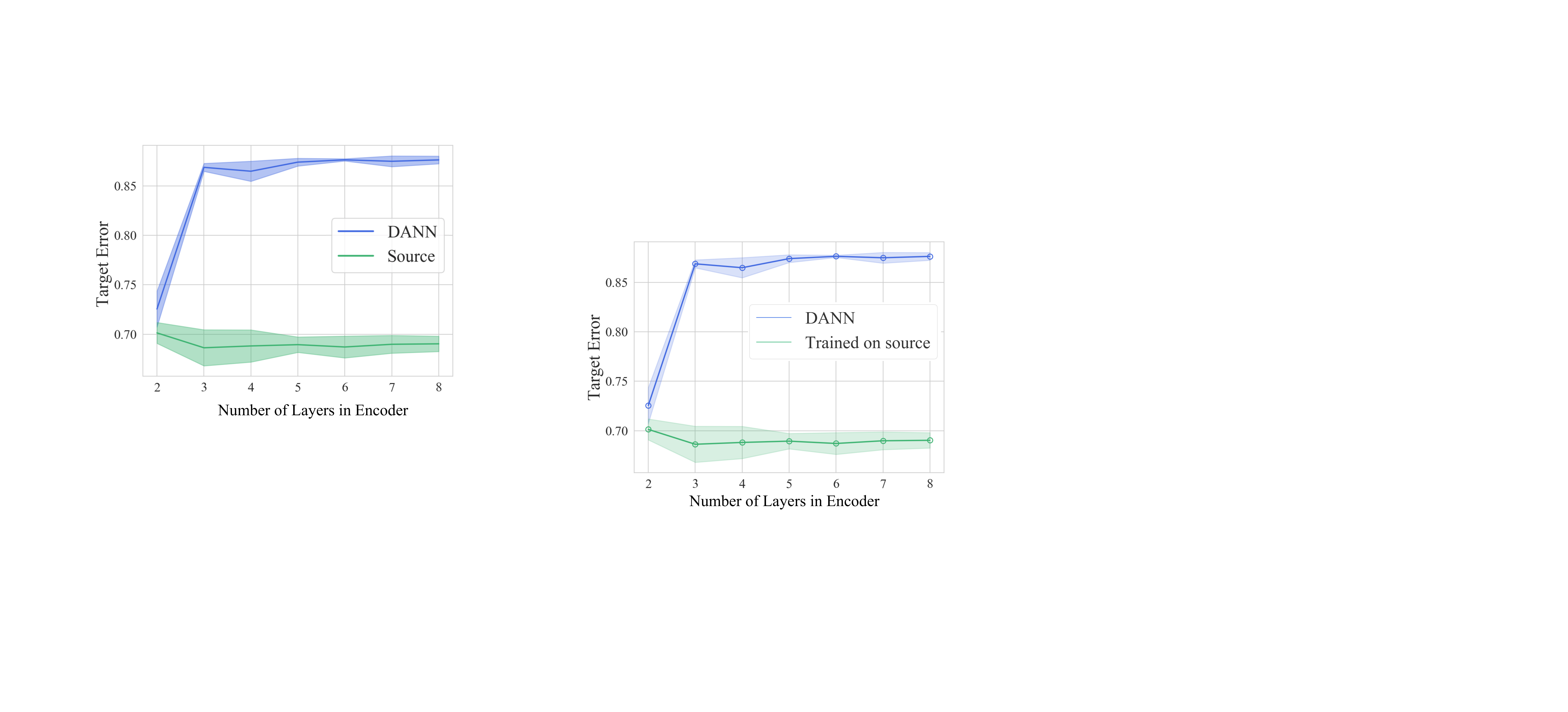}
  \end{center}
  \vspace{-5mm}
  \caption{DANN with FC layers.} \label{fig_ib}
  \vspace{-4mm}
\end{wrapfigure}

To analyze the effect of the embedding complexity, we augment the original two-layer CNN encoders with 1 to 6 additional CNN layers for M$\rightarrow$M-M and 1 to 24 for S$\rightarrow$M, leaving other settings unchanged. Figure \ref{fig_mnist}(a) shows the results. Again, the target error decreases initially and increase as the encoder becomes more complex. Notably, the target error increases by $19.8\%$ in M$\rightarrow$M-M and $8.8\%$ in S$\rightarrow$M compared to the optimal case, when more layers are added to the encoder. We also consider the width of hidden layers as a complexity measure,
%and multiply the hidden size of the encoder in the standard setting by a ratio
while fixing the depth of both encoder and predictor. The results are shown in Figure \ref{fig_mnist}(b). This time, the decrease in target error is not significant compared to increasing encoder depth. This suggests that depth plays a more important role than width in learning domain-invariant representations.

Next, we fix the total number of CNN layers of the neural network to $7$ and $26$ for M$\rightarrow$M-M and S$\rightarrow$M, respectively, and optimize the domain-invariant loss in different intermediate layers. The results in Figure \ref{fig_mnist}(c) again show a ``U-curve", indicating the existence of an optimal division.
%which can achieve significantly lower target error comparing to the others.
Even with fixed total size of the network ($\mathcal{H}$), the performance gap between different divisions can still reach $19.5\%$ in M$\rightarrow$M-M and $10.4\%$ in S$\rightarrow$M. For MDM, $\mathcal{L}$ contains all the augmented CNN layers for M$\rightarrow$M-M. For S$\rightarrow$M, we sub-sample a CNN layer every four layers to form $\mathcal{L}$. We also observe that MDM with all weighting schemes consistently achieves comparable performance with the best division in S$\rightarrow$M and even better performance in M$\rightarrow$M-M.

To investigate the importance of inductive bias in domain-invariant representations, we replace the CNN encoder by an MLP encoder. The results for M$\rightarrow$M-M are shown in Figure \ref{fig_ib}. Comparing to CNNs, which encode invariance via pooling and learned filters, MLPs do not have any inductive bias and lead to worse performance. In fact, the target error with MLP-based domain adaptation is higher than merely training on the source: without an appropriate inductive bias, learning domain invariant representations can even worsen the performance.

\paragraph{Object Classification.}

\begin{figure}
   
    \includegraphics[width=\linewidth]{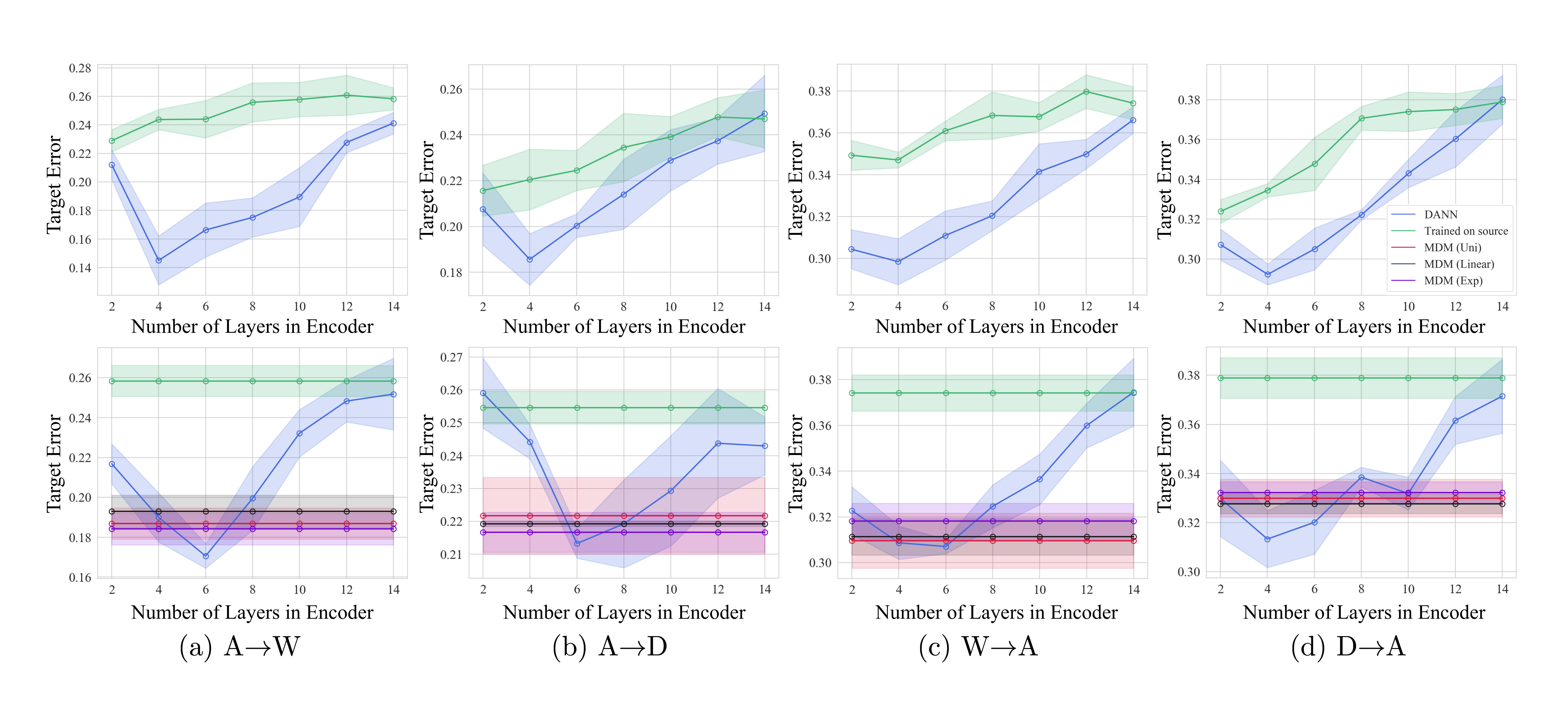}
    \vspace{-7mm}
    \caption{
       Office-31 Dataset. First row: Fixed predictor class, varying encoder depth. Second row: Fixed total number of layers, optimizing domain-invariant loss in a single layer or MDM.
    } \label{fig_office}
  \vspace{-1mm}
\end{figure}

Office-31 \citep{saenko2010adapting}, one of the most widely used benchmarks in domain adaptation, contains three domains: Amazon (A), Webcam (W), and DSLR (D) with 4,652 images and 31 categories. We show results for A$\rightarrow$W, A$\rightarrow$D, W$\rightarrow$A, and D$\rightarrow$A in Figure~\ref{fig_office}. To overcome the lack of training data, similar to \citep{li2018domain, long2018conditional}, we use ResNet-50 \citep{he2016deep} pretrained on ImageNet \citep{deng2009imagenet} for feature extraction. With the extracted features, we adopt multi-layer ReLU networks as hypothesis class. Again, we increase the depth of the encoder while fixing the depth of the predictor to $2$ and show the results Figure \ref{fig_office}.
Even with a powerful feature extractor, the embedding complexity tradeoff still exists.
Second, we fix the total network depth to 14 and optimize MDM, with $\mathcal{L}$ containing all even layers in the network.
%
%The second row of Figure \ref{fig_office} shows the target error when the domain-invariant loss is optimized in different layers while fixing the total number of layers to $14$. For MDM, the $\mathcal{L}$ contains all the even layers in the network.
MDM achieves comparable performance to the best division for most of the tasks, albeit slightly worse performance in D$\rightarrow$A.

\section{Conclusion}
\vspace{-2mm}
In this paper, we theoretically and empirically analyze the effect of embedding complexity on the target risk in domain-invariant representations. We find a complexity tradeoff that has mostly been overlooked by previous work. In fact, without carefully selecting and restricting the encoder class, learning domain invariant representations might even harm the performance. We further develop a simple yet effective algorithm to approximately optimize the tradeoff, achieving performance across tasks that matches the best network division, i.e., complexity tradeoff.
Interesting future directions of work include other strategies for model selection, and a more refined analysis and exploitation of the effect of inductive bias.

\vspace{-1mm}
\section*{Acknowledgements}
\vspace{-2mm}
This work was supported by MIT-IBM Watson AI Lab and NSF CAREER Award 1553284. We thank Tongzhou Wang, Joshua Robinson, Wei Fang, Wei-Chiu Ma, and Chen-Ming Chuang for helpful discussions and suggestions.

\bibliography{ms}
\bibliographystyle{ms}

\appendix
\section{Proofs}

\newtheorem{innercustomthm}{Theorem}
\newenvironment{customthm}[1]
  {\renewcommand\theinnercustomthm{#1}\innercustomthm}
  {\endinnercustomthm}

\subsection{Proof of Theorem 4}
\begin{customthm}{4}\label{four}
  For all $f \in \mathcal F$ and $g \in \mathcal G$,
\begin{align}
R_{T}(fg) \leq R_{S}(fg) + d_{\mathcal{F}\Delta\mathcal{F}}(p_{S}^{g}(Z), p_{T}^{g}(Z)) + d_{\mathcal{F}_{\mathcal{G}\Delta\mathcal{G}}}(p_{S}, p_{T})+\lambda_{\mathcal F\mathcal{G}}(g). \nonumber
\end{align}
where $\lambda_{\mathcal F\mathcal{G}}(g)$ is the best in-class joint risk defined as
\begin{align}
    \lambda_{\mathcal F\mathcal{G}}(g) =  \inf_{f^{\prime} \in \mathcal{F}, g^{\prime} \in \mathcal{G}} 2R_{S}(f^{\prime}g) + R_{S}(f^{\prime}g^{\prime}) + R_{T}(f^{\prime}g^{\prime}). \nonumber
\end{align}
\end{customthm}
\begin{proof}

We first define the optimal composition hypothesis $f^{\ast}g^{\ast}$ with respect to an encoder $g$ to be the hypothesis which minimizes the following error
\begin{align}
    f^{\ast}g^{\ast} = \argmin_{f^{\prime} \in \mathcal{F}, g^{\prime} \in \mathcal{G}} 2R_{S}(f^{\prime}g) + R_{S}(f^{\prime}g^{\prime}) + R_{T}(f^{\prime}g^{\prime})
\end{align}

By the triangle inequality for classification error (\citet{ben2007analysis}),
\begin{align}
R_{T}(fg) &\leq R_{T}(f^{\ast}g^{\ast}) + R_{T}(fg, f^{\ast}g^{\ast}) \\
&\leq R_{T}(f^{\ast}g^{\ast}) + R_{T}(fg, f^{\ast}g) + R_{T}(f^{\ast}g, f^{\ast}g^{\ast}) 
    \label{eq:1}
\end{align}
The second term on the right-hand side of Eq. \ref{eq:1} can be bounded as
\begin{align}
R_{T}(fg, f^{\ast}g) &\leq  R_{S}(fg, f^{\ast}g) + |R_{S}(fg, f^{\ast}g) - R_{T}(fg, f^{\ast}g)| \\
&\leq R_{S}(fg, f^{\ast}g) + \sup_{f,f^{\prime} \in \mathcal{F}}|R_{S}(fg, f^{\prime}g) - R_{T}(fg, f^{\prime}g)| \\
&= R_{S}(fg, f^{\ast}g) + d_{\mathcal{F}\Delta\mathcal{F}}(p_{S}^{g}(Z), p_{T}^{g}(Z)) \\
&\leq R_{S}(fg) + R_{S}(f^{\ast}g) + d_{\mathcal{F}\Delta\mathcal{F}}(p_{S}^{g}(Z), p_{T}^{g}(Z))
\end{align}
The third term on the right-hand side of Eq. \ref{eq:1} can be bounded as
\begin{align}
R_{T}(f^{\ast}g, f^{\ast}g^{\ast}) &\leq  R_{S}(f^{\ast}g, f^{\ast}g^{\ast}) + |R_{S}(f^{\ast}g, f^{\ast}g^{\ast}) - R_{T}(f^{\ast}g, f^{\ast}g^{\ast})| \\
&\leq R_{S}(f^{\ast}g, f^{\ast}g^{\ast}) + \sup_{f \in \mathcal{F}, g, g^{\prime} \in \mathcal{G} }|R_{S}(f^{\prime}g, f^{\prime}g^{\prime}) - R_{T}(f^{\prime}g, f^{\prime}g^{\prime})| \\
&= R_{S}(f^{\ast}g, f^{\ast}g^{\ast}) + d_{\mathcal{F}_{\mathcal{G}\Delta\mathcal{G}}}(p_{S}(X), p_{T}(X)) \\
&\leq  R_{S}(f^{\ast}g) + R_{S}(f^{\ast}g^{\ast}) + d_{\mathcal{F}_{\mathcal{G}\Delta\mathcal{G}}}(p_{S}(X), p_{T}(X))
\end{align}

Combine the above bounds, we have
\begin{align}
R_{T}(fg) &\leq R_{S}(fg) + d_{\mathcal{F}\Delta\mathcal{F}}(p_{S}^{g}(Z), p_{T}^{g}(Z)) + d_{\mathcal{F}_{\mathcal{G}\Delta\mathcal{G}}}(p_{S}(X), p_{T}(X)) + \lambda_{\mathcal{FG}}(g)
\end{align}
where
\begin{align}
\lambda_{\mathcal{FG}}(g) &= 2R_{S}(f^{\ast}g) + R_{S}(f^{\ast}g^{\ast}) + R_{T}(f^{\ast}g^{\ast}) \\
&= \inf_{f^{\prime} \in \mathcal{F}, g^{\prime} \in \mathcal{G}} 2R_{S}(f^{\prime}g) + R_{S}(f^{\prime}g^{\prime}) + R_{T}(f^{\prime}g^{\prime})
\end{align}
\end{proof}

\newtheorem{innercustomproposition}{Proposition}
\newenvironment{customproposition}[1]
  {\renewcommand\theinnercustomproposition{#1}\innercustomproposition}
  {\endinnercustomproposition}

\subsection{Proof of Proposition 5}

\begin{customproposition}{5}\label{five}
In an $N$-layer feedforward neural network $h = f_{i}g_{i} \in \mathcal{F}_{i}\mathcal{G}_{i} = \mathcal{H}$ for $i \in \{1, 2, \dots, N-1 \}$, the following inequalities hold for all $i \leq j$:
\begin{align}
    %d_{\mathcal{F}_{\mathcal{G}\Delta\mathcal{G}}}(p_{S}, p_{T}) \leq
    d_{{\mathcal{F}_{i}}_{\mathcal{G}_{i}\Delta\mathcal{G}_{i}}}(p_{S}, p_{T}) &\leq
    d_{{\mathcal{F}_{j}}_{\mathcal{G}_{j}\Delta\mathcal{G}_{j}}}(p_{S}, p_{T})  \nonumber \\
         d_{{\mathcal{F}_{i}}\Delta\mathcal{F}_{i}}(p_{S}^{g_{i}}(Z), p_{T}^{g_{i}}(Z)) &\geq
    d_{{\mathcal{F}_{j}}\Delta\mathcal{F}_{j}}(p_{S}^{g_{j}}(Z), p_{T}^{g_{j}}(Z)) \nonumber
\end{align}
\end{customproposition}

\begin{proof}
Given a class of multilayer feedforward neural network, We define a class of function $\mathcal{Q}_{ij}$ to represent the function class formed by the intermediate hidden layer $i$ to layer $j$. 

We now prove the first inequality. By the definition of $\mathcal{F}_{\mathcal{G}\Delta\mathcal{G}}$-divergence, for every $i \leq j$
\begin{align}
&d_{{\mathcal{F}_{i}}_{\mathcal{G}_{i}\Delta\mathcal{G}_{i}}}(p_{S}, p_{T}) \\
=& \sup_{\substack{f \in \mathcal{F}_{i} \\ g, g^{\prime} \in \mathcal{G}_{i}}} |R_{S}(fg, fg^{\prime}) - R_{T}(fg, fg^{\prime})| \\
=& \sup_{\substack{f \in \mathcal{F}_{j}, q \in \mathcal{Q}_{ij} \\ g, g^{\prime} \in \mathcal{G}_{i}}} |R_{S}(fqg, fqg^{\prime}) - R_{T}(fqg, fqg^{\prime})| \\
\leq& \sup_{\substack{f \in \mathcal{F}_{j} \\ q, q^{\prime} \in \mathcal{Q}_{ij} \\ g, g^{\prime} \in \mathcal{G}_{i}}} |R_{S}(fqg, fq^{\prime}g^{\prime}) - R_{T}(fqg, fq^{\prime}g^{\prime})| \\
=& \sup_{\substack{f \in \mathcal{F}_{j} \\ g, g^{\prime} \in \mathcal{G}_{j}}} |R_{S}(fg, fg^{\prime}) - R_{T}(fg, fg^{\prime})| \\
=&d_{{\mathcal{F}_{j}}_{\mathcal{G}_{j}\Delta\mathcal{G}_{j}}}(p_{S}, p_{T})
\end{align}

We next prove the second inequality. By the definition of $\mathcal{F}\Delta\mathcal{F}$-divergence, for every $i \leq j$
\begin{align}
&d_{\mathcal{F}_{j}\Delta\mathcal{F}_{j}}(p_{S}^{g_{j}}(Z), p_{T}^{g_{j}}(Z)) \\=& \sup_{f, f^{\prime} \in \mathcal{F}_{j}} |R_{S}(fg_{j}, f^{\prime}g_{j}) - R_{T}(fg_{j}, f^{\prime}g_{j})| \\
=& \sup_{f, f^{\prime} \in \mathcal{F}_{j}} |R_{S}(fq_{ij}g_{i}, f^{\prime}q_{ij}g_{i}) - R_{T}(fq_{ij}g_{i}, f^{\prime}q_{ij}g_{i})| \\
\leq&  \sup_{\substack{q \in \mathcal{Q}_{ij} \\ f, f^{\prime} \in \mathcal{F}_{j}}} |R_{S}(fqg_{i}, f^{\prime}qg_{i}) - R_{T}(fqg_{i}, f^{\prime}qg_{i})| \\
\leq&  \sup_{\substack{q, q^{\prime} \in \mathcal{Q}_{ij} \\ f, f^{\prime} \in \mathcal{F}_{j}}} |R_{S}(fqg_{i}, f^{\prime}q^{\prime}g_{i}) - R_{T}(fqg_{i}, f^{\prime}q^{\prime}g_{i})| \\
=& \sup_{f, f^{\prime} \in \mathcal{F}_{i}} |R_{S}(fg_{i}, f^{\prime}g_{i}) - R_{T}(fg_{i}, f^{\prime}g_{i})| \\
=& d_{\mathcal{F}_{i}\Delta\mathcal{F}_{i}}(p_{S}^{g_{i}}(Z), p_{T}^{g_{i}}(Z))
\end{align}
\end{proof}

\section{Predictor Complexity}
\label{a_pred_trad}

\begin{figure}[H]
    \includegraphics[width=\linewidth]{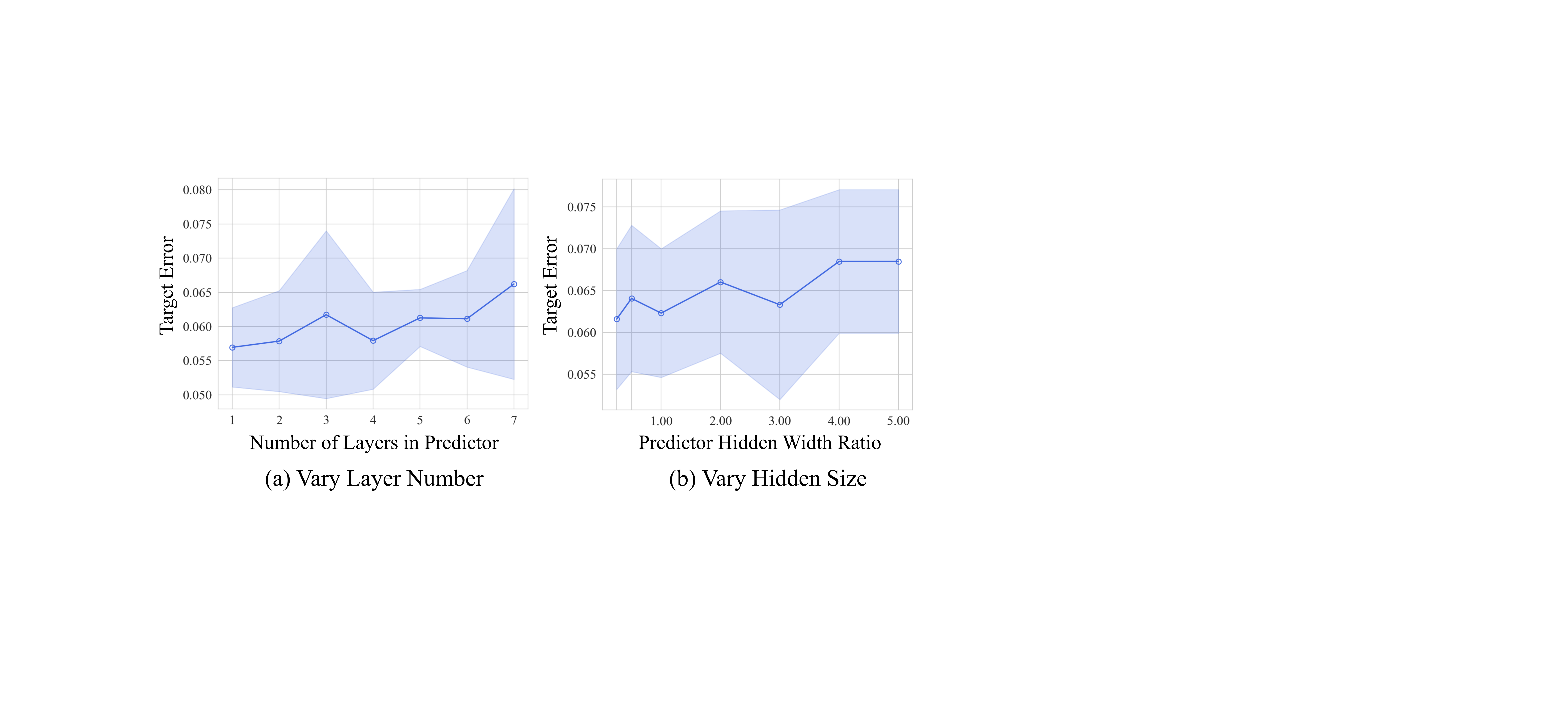}
    \vspace{-5mm}
    \caption{
       Predictor complexity trade-off on MNIST$\rightarrow$MNIST-M.  (a) Fix the encoder class and vary the number of layers in the predictor. (b) Fix the encoder class and vary the hidden width of the predictor.
    } \label{fig_pt}
  \vspace{-4mm}
\end{figure}

We investigate the effect of predictor complexity on MNIST$\rightarrow$MNIST-M. Follow the procedure in section \ref{exp}, we augment the original predictor with 1 to 7 additional CNN layers and fix the number of layers in encoder to $4$ or vary the hidden width. The results are shown in Figure \ref{fig_pt}. The target error slightly decreases as the number of layers in the predictor increases. Even we augment 7 layers to the predictor, the target error only decrease $0.9\%$ which is nearly ignorable. Therefore, we focus on the embedding complexity in the main paper which is both theoretically and empirically interesting.

\section{Experiment Details and Network Architectures}
\label{a_arch}

\subsection{Amazon Review Dataset}
\label{a_ard}
The learning rate of Adam optimizer is set to $1 \times e^{-3}$ and the model are trained for 50 epochs. We adopt the original progressive training strategy for discriminator \citep{ganin2016domain} where the weight $\alpha$ for domain-invariant loss in \eqref{di_obj} is initiated at $0$ and is gradually changed to $1$ using the following schedule:
\begin{align}
    \alpha = \frac{2}{1 + \exp(-10 \cdot p)} - 1
\end{align}
where $p$ is the training progress linearly changing from $0$ to $1$. The architecture of the hypothesis and discriminator are as follows:

\begin{center}
 \begin{tabular}{||c||} 
 \hline
 Encoder  \\ [0.5ex] 
 \hline\hline
 nn.Linear(5000, 128)  \\ 
 \hline
 nn.ReLU  \\
 \hline \hline
 nn.Linear(128, 128)  \\
 \hline
 nn.ReLU \\
 \hline
 $\times n$ (depends on the number of layers) \\
 \hline
\end{tabular}
\quad
 \begin{tabular}{||c||} 
 \hline
 Predictor  \\ [0.5ex] 
 \hline\hline
 nn.Linear(128, 128)  \\ 
 \hline
 nn.ReLU  \\
 \hline
 $\times n$ (depends on the number of layers) \\
 \hline \hline
 nn.Linear(128, 2)  \\
 \hline
 nn.Softmax \\
 \hline
\end{tabular}
\end{center}

\begin{center}
 \begin{tabular}{||c||} 
 \hline
 Discriminator  \\ [0.5ex] 
 \hline\hline
 nn.Linear(128, 256)  \\ 
 \hline
 nn.ReLU  \\
 \hline\hline
  nn.Linear(256, 256)  \\ 
 \hline
 nn.ReLU  \\
 \hline
 $\times$5\\
 \hline\hline
 nn.Linear(256, 2)  \\
 \hline
 nn.Softmax \\
 \hline
\end{tabular}
\end{center}

\subsection{Digit Classification}
The learning rate of Adam optimizer is set to $1 \times e^{-3}$ and the model are trained for 100 epochs. The weight $\alpha$ for domain-invariant loss in \eqref{di_obj} is initiated at $0$ and is gradually changed to $0.1$ using the same schedule in section \ref{a_ard}. The architecture of the hypothesis and discriminator are as follows:

\begin{center}
 \begin{tabular}{||c||} 
 \hline
 Encoder  \\ [0.5ex] 
 \hline\hline
 nn.Conv2d(3, 64, kernel$\_$size=5)  \\ 
 \hline
 nn.BatchNorm2d \\
 \hline
 nn.MaxPool2d(2)  \\
 \hline 
 nn.ReLU \\
 \hline
 nn.Conv2d(64, 128, kernel$\_$size=5)  \\ 
 \hline
 nn.BatchNorm2d \\
 \hline
 nn.Dropout2d (only added for MNIST$\rightarrow$MNIST-M)\\
 \hline
 nn.MaxPool2d(2)  \\
 \hline 
 nn.ReLU \\
 \hline \hline
 nn.Conv2d(128, 128, kernel$\_$size=3, padding=1)  \\
 \hline
 nn.BatchNorm2d \\
 \hline
 nn.ReLU \\
 \hline
 $\times n$ (depends on the number of layers) \\
 \hline
\end{tabular}
\end{center}

\begin{center}
 \begin{tabular}{||c||} 
 \hline
 Predictor  \\ [0.5ex] 
 \hline\hline
 nn.Conv2d(128, 128, kernel$\_$size=3, padding=1)  \\
 \hline
 nn.BatchNorm2d \\
 \hline
 nn.ReLU \\
 \hline
 $\times n$ (depends on the number of layers) \\
 \hline\hline
 flatten \\
 \hline
 nn.Linear(2048, 256)  \\ 
 \hline 
 nn.BatchNorm1d \\
 \hline
 nn.ReLU  \\
 \hline
 nn.Linear(256, 10) \\
 \hline 
 nn.Softmax \\
 \hline
\end{tabular}
\end{center}

\begin{center}
 \begin{tabular}{||c||} 
 \hline
 Discriminator  \\ [0.5ex] 
 \hline\hline
 nn.Conv2d(128, 256, kernel$\_$size=3, padding=1)  \\
 \hline
 nn.ReLU \\
 \hline
 \hline\hline
 nn.Conv2d(256, 256, kernel$\_$size=3, padding=1)  \\
 \hline
 nn.ReLU \\
 \hline
 $\times 4$ \\
 \hline\hline
 Flatten \\
 \hline
 nn.Linear(4096, 512)  \\ 
 \hline 
 nn.ReLU  \\
 \hline
 nn.Linear(512, 512)  \\ 
 \hline 
 nn.ReLU  \\
 \hline 
 nn.Linear(512, 2) \\
 \hline 
 nn.Softmax \\
 \hline
\end{tabular}
\end{center}

In the hidden width experiments, we treat the architectures above as the pivot and multiply their hidden width with the ratios.

\subsection{Office-31}
We exploit the feature after average pooling layer of the ResNet-50 \citep{he2016deep} pretrained on ImageNet \citep{deng2009imagenet} for feature extraction.
The learning rate of Adam optimizer is set to $3 \times e^{-4}$ and the model are trained for 100 epochs. The weight $\alpha$ for domain-invariant loss in \eqref{di_obj} is initiated at $0$ and is gradually changed to $1$ using the same schedule in section \ref{a_ard}. The architecture of the hypothesis and discriminator are as follows:

\begin{center}
 \begin{tabular}{||c||} 
 \hline
 Encoder  \\ [0.5ex] 
 \hline\hline
 nn.Linear(2048, 256)  \\ 
 \hline
 nn.BatchNorm1d \\
 \hline
 nn.ReLU  \\
 \hline \hline
 nn.Linear(256, 256)  \\
 \hline
 nn.BatchNorm1d \\
 \hline
 nn.ReLU \\
 \hline
 $\times n$ (depends on the number of layers) \\
 \hline
\end{tabular}
\quad
 \begin{tabular}{||c||} 
 \hline
 Predictor  \\ [0.5ex] 
 \hline\hline
 nn.Linear(256, 256)  \\ 
 \hline
 nn.BatchNorm1d \\
 \hline
 nn.ReLU  \\
 \hline
 $\times n$ (depends on the number of layers) \\
 \hline \hline
 nn.Linear(256, 2)  \\
 \hline
 nn.Softmax \\
 \hline
\end{tabular}
\end{center}

\begin{center}
 \begin{tabular}{||c||} 
 \hline
 Discriminator  \\ [0.5ex] 
 \hline\hline
  nn.Linear(256, 256)  \\ 
 \hline
 nn.ReLU  \\
 \hline
 $\times$6\\
 \hline\hline
 nn.Linear(256, 2)  \\
 \hline
 nn.Softmax \\
 \hline
\end{tabular}
\end{center}

\subsection{Multilayer Divergence Minimization}
In all the experiments, we minimize the divergence in multiple layers by augmenting additional discriminators for each layer-specific representations where the discriminators share the same architecture as the standard setting. 

For uniform weighting scheme ($\alpha_{i} = \alpha_{0}$), $\alpha_i$ is set to the normalized same value $\alpha$ in the stand setting. For linear decreasing scheme ($\alpha_{i} = \alpha_{0} - c \times i$), $\alpha_i$ decreases from $\alpha_{0}=\alpha$ to $0$ linearly. For exponentially decreasing scheme ($\alpha_{i} = \alpha_{0}\exp(-c \times i)$), $\alpha_0$ is set to $\alpha$ and $c$ increases from $0$ to $2$ linearly.

\end{document}

%% file: math_commands.tex
\usepackage{amsmath,amsfonts,bm}

\def\eqreff#1{(\ref{#1})}
\def\eqref#1{equation~(\ref{#1})}
\def\1{\bm{1}}

\DeclareMathAlphabet{\mathsfit}{\encodingdefault}{\sfdefault}{m}{sl}
\SetMathAlphabet{\mathsfit}{bold}{\encodingdefault}{\sfdefault}{bx}{n}

\DeclareMathOperator*{\argmin}{arg\,min}

%% file: ms.bbl
\begin{thebibliography}{28}
\providecommand{\natexlab}[1]{#1}
\providecommand{\url}[1]{\texttt{#1}}
\expandafter\ifx\csname urlstyle\endcsname\relax
  \providecommand{\doi}[1]{doi: #1}\else
  \providecommand{\doi}{doi: \begingroup \urlstyle{rm}\Url}\fi

\bibitem[Ben-David et~al.(2007)Ben-David, Blitzer, Crammer, and
  Pereira]{ben2007analysis}
Shai Ben-David, John Blitzer, Koby Crammer, and Fernando Pereira.
\newblock Analysis of representations for domain adaptation.
\newblock In \emph{Advances in neural information processing systems}, pp.\
  137--144, 2007.

\bibitem[Ben-David et~al.(2010)Ben-David, Blitzer, Crammer, Kulesza, Pereira,
  and Vaughan]{ben2010theory}
Shai Ben-David, John Blitzer, Koby Crammer, Alex Kulesza, Fernando Pereira, and
  Jennifer~Wortman Vaughan.
\newblock A theory of learning from different domains.
\newblock \emph{Machine learning}, 79\penalty0 (1-2):\penalty0 151--175, 2010.

\bibitem[Bousmalis et~al.(2016)Bousmalis, Trigeorgis, Silberman, Krishnan, and
  Erhan]{bousmalis2016domain}
Konstantinos Bousmalis, George Trigeorgis, Nathan Silberman, Dilip Krishnan,
  and Dumitru Erhan.
\newblock Domain separation networks.
\newblock In \emph{Advances in neural information processing systems}, pp.\
  343--351, 2016.

\bibitem[Cortes \& Mohri(2011)Cortes and Mohri]{cortes2011domain}
Corinna Cortes and Mehryar Mohri.
\newblock Domain adaptation in regression.
\newblock In \emph{International Conference on Algorithmic Learning Theory},
  pp.\  308--323. Springer, 2011.

\bibitem[Courty et~al.(2016)Courty, Flamary, Tuia, and
  Rakotomamonjy]{courty2016optimal}
Nicolas Courty, R{\'e}mi Flamary, Devis Tuia, and Alain Rakotomamonjy.
\newblock Optimal transport for domain adaptation.
\newblock \emph{IEEE transactions on pattern analysis and machine
  intelligence}, 39\penalty0 (9):\penalty0 1853--1865, 2016.

\bibitem[Courty et~al.(2017)Courty, Flamary, Habrard, and
  Rakotomamonjy]{courty2017joint}
Nicolas Courty, R{\'e}mi Flamary, Amaury Habrard, and Alain Rakotomamonjy.
\newblock Joint distribution optimal transportation for domain adaptation.
\newblock In \emph{Advances in Neural Information Processing Systems}, pp.\
  3730--3739, 2017.

\bibitem[Deng et~al.(2009)Deng, Dong, Socher, Li, Li, and
  Fei-Fei]{deng2009imagenet}
Jia Deng, Wei Dong, Richard Socher, Li-Jia Li, Kai Li, and Li~Fei-Fei.
\newblock Imagenet: A large-scale hierarchical image database.
\newblock In \emph{2009 IEEE conference on computer vision and pattern
  recognition}, pp.\  248--255. Ieee, 2009.

\bibitem[Ganin \& Lempitsky(2014)Ganin and Lempitsky]{ganin2014unsupervised}
Yaroslav Ganin and Victor Lempitsky.
\newblock Unsupervised domain adaptation by backpropagation.
\newblock \emph{arXiv preprint arXiv:1409.7495}, 2014.

\bibitem[Ganin et~al.(2016)Ganin, Ustinova, Ajakan, Germain, Larochelle,
  Laviolette, Marchand, and Lempitsky]{ganin2016domain}
Yaroslav Ganin, Evgeniya Ustinova, Hana Ajakan, Pascal Germain, Hugo
  Larochelle, Fran{\c{c}}ois Laviolette, Mario Marchand, and Victor Lempitsky.
\newblock Domain-adversarial training of neural networks.
\newblock \emph{The Journal of Machine Learning Research}, 17\penalty0
  (1):\penalty0 2096--2030, 2016.

\bibitem[He et~al.(2016)He, Zhang, Ren, and Sun]{he2016deep}
Kaiming He, Xiangyu Zhang, Shaoqing Ren, and Jian Sun.
\newblock Deep residual learning for image recognition.
\newblock In \emph{Proceedings of the IEEE conference on computer vision and
  pattern recognition}, pp.\  770--778, 2016.

\bibitem[Hoffman et~al.(2017)Hoffman, Tzeng, Park, Zhu, Isola, Saenko, Efros,
  and Darrell]{hoffman2017cycada}
Judy Hoffman, Eric Tzeng, Taesung Park, Jun-Yan Zhu, Phillip Isola, Kate
  Saenko, Alexei~A Efros, and Trevor Darrell.
\newblock Cycada: Cycle-consistent adversarial domain adaptation.
\newblock \emph{arXiv preprint arXiv:1711.03213}, 2017.

\bibitem[Johansson et~al.(2019)Johansson, Ranganath, and
  Sontag]{johansson2019support}
Fredrik~D Johansson, Rajesh Ranganath, and David Sontag.
\newblock Support and invertibility in domain-invariant representations.
\newblock \emph{arXiv preprint arXiv:1903.03448}, 2019.

\bibitem[Kingma \& Ba(2014)Kingma and Ba]{kingma2014adam}
Diederik~P Kingma and Jimmy Ba.
\newblock Adam: A method for stochastic optimization.
\newblock \emph{arXiv preprint arXiv:1412.6980}, 2014.

\bibitem[Lee \& Raginsky(2018)Lee and Raginsky]{lee2018minimax}
Jaeho Lee and Maxim Raginsky.
\newblock Minimax statistical learning with wasserstein distances.
\newblock In \emph{Advances in Neural Information Processing Systems}, pp.\
  2687--2696, 2018.

\bibitem[Li et~al.(2018)Li, Song, Huang, Ding, and Wu]{li2018domain}
Shuang Li, Shiji Song, Gao Huang, Zhengming Ding, and Cheng Wu.
\newblock Domain invariant and class discriminative feature learning for visual
  domain adaptation.
\newblock \emph{IEEE Transactions on Image Processing}, 27\penalty0
  (9):\penalty0 4260--4273, 2018.

\bibitem[Long et~al.(2014)Long, Wang, Ding, Sun, and Yu]{long2014transfer}
Mingsheng Long, Jianmin Wang, Guiguang Ding, Jiaguang Sun, and Philip~S Yu.
\newblock Transfer joint matching for unsupervised domain adaptation.
\newblock In \emph{Proceedings of the IEEE conference on computer vision and
  pattern recognition}, pp.\  1410--1417, 2014.

\bibitem[Long et~al.(2015)Long, Cao, Wang, and Jordan]{long2015learning}
Mingsheng Long, Yue Cao, Jianmin Wang, and Michael~I Jordan.
\newblock Learning transferable features with deep adaptation networks.
\newblock \emph{arXiv preprint arXiv:1502.02791}, 2015.

\bibitem[Long et~al.(2016)Long, Zhu, Wang, and Jordan]{long2016unsupervised}
Mingsheng Long, Han Zhu, Jianmin Wang, and Michael~I Jordan.
\newblock Unsupervised domain adaptation with residual transfer networks.
\newblock In \emph{Advances in Neural Information Processing Systems}, pp.\
  136--144, 2016.

\bibitem[Long et~al.(2017)Long, Zhu, Wang, and Jordan]{long2017deep}
Mingsheng Long, Han Zhu, Jianmin Wang, and Michael~I Jordan.
\newblock Deep transfer learning with joint adaptation networks.
\newblock In \emph{Proceedings of the 34th International Conference on Machine
  Learning-Volume 70}, pp.\  2208--2217. JMLR. org, 2017.

\bibitem[Long et~al.(2018)Long, Cao, Wang, and Jordan]{long2018conditional}
Mingsheng Long, Zhangjie Cao, Jianmin Wang, and Michael~I Jordan.
\newblock Conditional adversarial domain adaptation.
\newblock In \emph{Advances in Neural Information Processing Systems}, pp.\
  1640--1650, 2018.

\bibitem[Mansour et~al.(2009)Mansour, Mohri, and
  Rostamizadeh]{mansour2009domain}
Yishay Mansour, Mehryar Mohri, and Afshin Rostamizadeh.
\newblock Domain adaptation: Learning bounds and algorithms.
\newblock \emph{arXiv preprint arXiv:0902.3430}, 2009.

\bibitem[Saenko et~al.(2010)Saenko, Kulis, Fritz, and
  Darrell]{saenko2010adapting}
Kate Saenko, Brian Kulis, Mario Fritz, and Trevor Darrell.
\newblock Adapting visual category models to new domains.
\newblock In \emph{European conference on computer vision}, pp.\  213--226.
  Springer, 2010.

\bibitem[Shen et~al.(2017)Shen, Qu, Zhang, and Yu]{shen2017wasserstein}
Jian Shen, Yanru Qu, Weinan Zhang, and Yong Yu.
\newblock Wasserstein distance guided representation learning for domain
  adaptation.
\newblock \emph{arXiv preprint arXiv:1707.01217}, 2017.

\bibitem[Shu et~al.(2018)Shu, Bui, Narui, and Ermon]{shu2018dirt}
Rui Shu, Hung~H Bui, Hirokazu Narui, and Stefano Ermon.
\newblock A dirt-t approach to unsupervised domain adaptation.
\newblock \emph{arXiv preprint arXiv:1802.08735}, 2018.

\bibitem[Tzeng et~al.(2015)Tzeng, Hoffman, Darrell, and
  Saenko]{tzeng2015simultaneous}
Eric Tzeng, Judy Hoffman, Trevor Darrell, and Kate Saenko.
\newblock Simultaneous deep transfer across domains and tasks.
\newblock In \emph{Proceedings of the IEEE International Conference on Computer
  Vision}, pp.\  4068--4076, 2015.

\bibitem[Tzeng et~al.(2017)Tzeng, Hoffman, Saenko, and
  Darrell]{tzeng2017adversarial}
Eric Tzeng, Judy Hoffman, Kate Saenko, and Trevor Darrell.
\newblock Adversarial discriminative domain adaptation.
\newblock In \emph{Proceedings of the IEEE Conference on Computer Vision and
  Pattern Recognition}, pp.\  7167--7176, 2017.

\bibitem[Wu et~al.(2019)Wu, Winston, Kaushik, and Lipton]{wu2019domain}
Yifan Wu, Ezra Winston, Divyansh Kaushik, and Zachary Lipton.
\newblock Domain adaptation with asymmetrically-relaxed distribution alignment.
\newblock \emph{arXiv preprint arXiv:1903.01689}, 2019.

\bibitem[Zhao et~al.(2019)Zhao, Combes, Zhang, and Gordon]{zhao2019learning}
Han Zhao, Remi Tachet~des Combes, Kun Zhang, and Geoffrey~J Gordon.
\newblock On learning invariant representation for domain adaptation.
\newblock \emph{arXiv preprint arXiv:1901.09453}, 2019.

\end{thebibliography}
